\documentclass{article} 
\usepackage{iclr2027_arxiv_preprint,times}

\usepackage{amsmath,amsfonts,bm}

\def\eqref#1{equation~\ref{#1}}

\def\1{\bm{1}}

\DeclareMathAlphabet{\mathsfit}{\encodingdefault}{\sfdefault}{m}{sl}
\SetMathAlphabet{\mathsfit}{bold}{\encodingdefault}{\sfdefault}{bx}{n}

\usepackage{hyperref}
\usepackage{url}

\usepackage{silence}
\usepackage[utf8]{inputenc}
\usepackage[T1]{fontenc}
\usepackage{tabularx}
\usepackage{hyperref}
\usepackage{lipsum}
\usepackage{url}
\usepackage{placeins}
\usepackage{needspace}
\usepackage{booktabs}
\usepackage{soul}
\usepackage{amsfonts}
\usepackage{nicefrac}
\usepackage{microtype}
\usepackage{natbib}
\usepackage{proba}
\usepackage{wrapfig}

\usepackage{caption}
\usepackage{makecell}
\usepackage{amssymb}
\usepackage{pifont}

\usepackage{subcaption}
\usepackage{float}
\usepackage{rotating}
\usepackage{etoolbox}
\usepackage{xparse}
\usepackage{grffile}
\usepackage{amsmath}
\usepackage{xspace}
\usepackage{euscript}
\usepackage{enumitem}
\usepackage{mathtools}
\usepackage{tikz}
\usepackage{tablefootnote}
\usepackage[flushleft]{threeparttable}
\usepackage{multirow}
\usepackage[title]{appendix}
\usepackage{arydshln}
\usepackage{array, booktabs}
\usepackage{comment}
\usepackage{graphicx}
\usepackage{adjustbox}
\usepackage{dsfont}
\usepackage{amsmath,amsthm}
\usepackage{balance}
\usepackage{multicol}
\usepackage{nameref}
\usepackage{pdfpages}
\usepackage{braket}
\usepackage[normalem]{ulem}
\usepackage{bbding}
\usepackage[capitalize,noabbrev]{cleveref}
\crefname{appendix}{Appendix}{Appendices}
\Crefname{appendix}{Appendix}{Appendices}
\usepackage{xfrac}
\usepackage[usestackEOL]{stackengine}
\usepackage{indentfirst}
\usepackage{csquotes}
\usepackage{pgf}
\usepackage{varwidth}
\usepackage{verbatimbox}
\usepackage{verbatim}
\usepackage{bm}
\usepackage{apptools}
\usepackage[ruled,vlined,linesnumbered]{algorithm2e}
\usepackage[colorinlistoftodos,prependcaption,textsize=small]{todonotes}

\usepackage{empheq}
\definecolor{myblue}{rgb}{.8, .8, 1}

\definecolor{pastelblue}{RGB}{76,113,175}
\definecolor{pastelgreen}{RGB}{144,238,144}
\definecolor{pastelred}{RGB}{196,78,82}
\definecolor{pastelgrey}{RGB}{230,230,230}
\definecolor{pastelbeige}{RGB}{243,236,221}
\definecolor{pastelpurple}{RGB}{154,139,192}
\definecolor{salmon}{RGB}{250, 128, 114}
\definecolor{darkgreen}{rgb}{0,0.6,0}
\definecolor{darkred}{rgb}{0.5,0,0}
\definecolor{verylightgreen}{HTML}{F6FFF9}
\definecolor{verylightred}{HTML}{FFF4F3}
\definecolor{verylightgray}{HTML}{F4F6F6}
\definecolor{babyblueeyes}{rgb}{0.63, 0.79, 0.95}
\definecolor{lightpink}{rgb}{1.00, 0.714, 0.757}

\usepackage{tikzpeople}
\usetikzlibrary{shapes,decorations,arrows,calc,arrows.meta,fit,positioning}

\tikzset{
    -Latex,auto,node distance =1 cm and 1 cm,semithick,
    state/.style ={ellipse, draw, minimum width = 0.7 cm},
    point/.style = {circle, draw, inner sep=0.04cm,fill,node contents={}},
    bidirected/.style={Latex-Latex,dashed},
    el/.style = {inner sep=2pt, align=left, sloped}
}

 \newtheorem{proposition}{Proposition}
 \theoremstyle{remark}
 
\makeatletter
\def\thmt@refnamewithcomma #1#2#3,#4,#5\@nil{%
    \@xa\def\csname\thmt@envname #1utorefname\endcsname{#3}%
    \ifcsname #2refname\endcsname
        \csname #2refname\expandafter\endcsname\expandafter{\thmt@envname}{#3}{#4}%
    \fi}
\makeatother
\Crefname{conjecture}{Conjecture}{Conjectures}
\Crefname{definition}{Definition}{Definitions}
\Crefname{observation}{Observation}{Observations}
\Crefname{assumption}{Assumption}{Assumptions}
\Crefname{axiom}{Axiom}{Axioms}
\Crefname{case}{Case}{Cases}
\Crefname{claim}{Claim}{Claims}
\Crefname{conclusion}{Conclusion}{Conclusions}
\Crefname{condition}{Condition}{Conditions}
\Crefname{criterion}{Criterion}{Criteria}
\Crefname{exercise}{Exercise}{Exercises}
\Crefname{example}{Example}{Examples}
\Crefname{notation}{Notation}{Notations}
\Crefname{problem}{Problem}{Problems}
\Crefname{property}{Property}{Properties}
\Crefname{remark}{Remark}{Remarks}
\Crefname{solution}{Solution}{Solutions}
\Crefname{summary}{Summary}{Summaries}
\Crefname{motivation}{Motivation}{Motivations}
\Crefname{query}{Query}{Queries}
\crefname{algocf}{Alg.}{Algs.}
\Crefname{algocf}{Algorithm}{Algorithms}

\newcommand{\blue}[1]{{\color{blue} #1}}
\newcommand*\dbar[1]{\overline{\overline{\lower0.2ex\hbox{$#1$}}}}

\DeclareFontFamily{U}{BOONDOX-calo}{\skewchar\font=45 }
\DeclareFontShape{U}{BOONDOX-calo}{m}{n}{
    <-> s*[1.05] BOONDOX-r-calo}{}
\DeclareFontShape{U}{BOONDOX-calo}{b}{n}{
    <-> s*[1.05] BOONDOX-b-calo}{}
\DeclareMathAlphabet{\mathcalb}{U}{BOONDOX-calo}{m}{n}
\SetMathAlphabet{\mathcalb}{bold}{U}{BOONDOX-calo}{b}{n}
\DeclareMathAlphabet{\mathbcalb}{U}{BOONDOX-calo}{b}{n}

\usepackage{cancel}
\renewcommand{\paragraph}[1]{{\noindent \textbf{#1.}}}

\definecolor{pastel_purple}{HTML}{756FB3}
\definecolor{pastel_green}{HTML}{1D9D79}

\definecolor{greenfkc}{rgb}{0.235000,0.567222,0.524073}

\colorlet{PastelPurpleLight}{pastel_purple!15!white}
\colorlet{PastelGreenLight}{pastel_green!15!white}
\sethlcolor{PastelPurpleLight}
\sethlcolor{PastelGreenLight}

\usepackage[many]{tcolorbox}

\newtcbox{\addedbox}[1][PastelGreenLight]{
    on line,
    arc=1pt,
    colback={#1},
    colframe={#1},
    boxrule=0pt,
    boxsep=0pt,
    left=0pt,
    right=0pt,
    top=0pt,
    bottom=0pt
}
\definecolor{customteal}{HTML}{007A87}

\definecolor{highlightblue}{HTML}{BDD9DB}
\definecolor{highlightpink}{HTML}{E5A1B0}

\definecolor{highlightyellow}{HTML}{FFDFB3}

\title{Adaptive Order Policies for Masked Diffusion}

\author{Jama Hussein Mohamud$^{1,2}$\thanks{Equal contribution.\newline \hspace*{1.8em} Correspondence to ~\texttt{\{hussein-mohamu.jama, mohsin.hasan\}@mila.quebec}}~,~~ Mohsin Hasan$^{1,2}$\footnotemark[1]~,~~ Mirco Ravanelli$^{2,3}$,
\bf{Yoshua Bengio}$^{1, 2, 4}$ \vspace{1em} \\
$^1$Universit\'e de Montr\'eal, $^2$Mila, $^3$Concordia University, $^4$LawZero
}

\iclrfinalcopy 

\begin{document}

\maketitle
\begin{abstract}
    Masked diffusion models have seen great success in capturing data distributions over discrete sequences in domains such as text and proteins. These models generate data by iteratively unmasking tokens starting from a fully masked sequence, with the unmasking order typically chosen at random or using a heuristic based on denoiser probabilities. In this work, we propose a scheme for learning the unmasking order using an additional lightweight policy network on top of a diffusion model. Our proposed loss reweights terms in the masked diffusion loss according to policy probabilities, and results in a policy that prefers positions where the denoiser is more likely to be correct. We study this loss in two settings: (i) training solely the policy while using a frozen pre-trained denoiser, and (ii) training the policy and denoiser jointly with the weighted loss to allow for mutual adaptation. We demonstrate that our approach outperforms common heuristics on problems that are sensitive to token ordering, such as combinatorial problems, proteins as well as various coding and language tasks.
\end{abstract}

\section{Introduction}
\label{sec:introduction}

Diffusion models have established themselves as a powerful paradigm for generative modeling, achieving remarkable success in continuous domains such as images \citep{ho2020denoisingdiffusionprobabilisticmodels, saharia2022photorealistic, Rombach_2022_CVPR} and molecular structures \citep{watson2023novo, abramson2024accurate}. More recently, \emph{discrete} diffusion models -- which operate directly on token sequences by iteratively masking and unmasking -- have shown strong results in language modeling \citep{sahoo2024simple, nie2025llada, shi2024simplified}, protein design \citep{alamdari2023protein, wang2024dplm}, and drug discovery \citep{lee2025genmol}. \citep{huang2022riemannian}\citep{chen2024flow}\citep{austin2021structured} \citep{gat2024discrete}

A key design choice in masked diffusion models (MDM) is the \emph{order} in which tokens are unmasked during generation. The standard approach selects positions uniformly at random. However, practitioners have found that heuristic ordering strategies -- such as unmasking the most confident position first \citep{nie2025llada} or the position with the largest probability margin \citep{kim2025trainworstplanbest} -- can dramatically improve sample quality on downstream tasks. This effect is particularly pronounced on constraint satisfaction problems such as Sudoku and Boolean satisfiability (3-SAT), where the unmasking order directly impacts whether the model can propagate constraints correctly. 


Despite the empirical success of heuristic orderings, these remain hand-designed and may be suboptimal for a given model and dataset. A natural question arises: \emph{can we learn the unmasking order?} That is, rather than relying on fixed heuristics, can we train a lightweight auxiliary network to predict which positions to unmask, conditioned on the current partially masked sequence?

In this work, we propose a simple approach for learning adaptive unmasking orderings in MDMs. Our approach introduces a policy network $q^\phi(i \mid x_t)$ and a modified cross-entropy objective that can be used either to train a lightweight policy layer on top of a pretrained masked diffusion model or to jointly train the policy and denoiser. The objective weights policy probabilities by the cross entropy of the denoiser at each token position, encouraging the policy to select positions that are most informative for generation. Under joint training, the objective also optimizes the denoiser to more correctly predict positions which the policy is likely to unmask. Across structured tasks, such as Sudoku, 3-SAT, and DPLM protein generation, learned ordering improves over heuristic baselines in both policy-only and joint-training settings. Policy-only adaptation also improves performance in larger-scale settings, such as with Open-dLLM-0.5B and LLaDA-8B across code, language, and reasoning tasks. 
The policy is lightweight, adding less than $1\%$ to the denoiser's parameter count and requiring far fewer training updates than the base model. In addition to performance over baselines, we observe a number of benefits of the learned policy over heuristics, including: i) robustness to stochastic decoding, ii) improved efficiency at inference - retaining higher performance at fewer diffusion steps, iii) improved heuristic decoding when the denoiser is trained with policy-aware weighting.

\section{Method}
\label{sec:method}

\subsection{Masked Diffusion Models}

Throughout this work, we denote a sequence of length $L$ as $x = (x^1, \dots, x^L) \in \mathcal{V}^L$, with tokens taking values in some vocabulary set $x^i \in \mathcal{V}$. We consider the case of masked diffusion, where a special masking token $m$ is included in the vocabulary set. Other notation includes: the Kronecker symbol $\delta(i, j)$ (equal to 1 for $i=j$ and 0 otherwise), $\mathrm{Cat}(x; p)$ to denote the categorical distribution with probabilities $p$, and $\Delta^k$ to denote the probability simplex over $k$ dimensions.

Masked diffusion models (MDMs) use a noising process to map the data distribution $p_\mathrm{data}(x)$ at time $0$ to the delta distribution at the fully masked state $M = (m,\dots,m)$, $p_1(x) = \delta(x,M)$ at time $1$. A typical noising process consists of converting a data token $x^i_0$ into the masked token with some probability $1-\alpha_t$, independently over dimensions \citep{sahoo2024simple}: $p(x_t \mid x_0) = \prod^L_{i=1} \alpha_t \delta(x^i_t, x^i_0) + (1- \alpha_t) \delta(x^i_t,m)$. The parameter $\alpha_t$ denotes a decreasing noise schedule, with $\alpha_0 = 1$ and $\alpha_1 = 0$. A typical choice is the linear schedule $\alpha_t = 1-t$.

For reversing this process, a neural network parameterizes a distribution over clean data $x_0$ conditioned on the partially masked sequence $x_t$. In particular, the network outputs an independent distribution over each token position $i$, as $\mu^\theta(x_t)[i, \cdot] \in \Delta^{|\mathcal{V}|}$, which satisfies $\mu^\theta(x_t)[i, m] = 0$ (the clean data cannot contain masks) and $\mu^\theta(x_t)[i, x^i_t] = 1$ if $x^i_t \neq m$ (the clean data approximation retains unmasked positions in $x_t$). The function $\mu^\theta$ is referred to as the denoiser.

Given a denoiser, the reverse transition over two nearby time-steps $s < t$ is \citep{sahoo2024simple}:
\begin{align}
    \label{eq:reverse_transition}
    p^\theta(x^i_{s} \mid x_t) = \begin{cases}
                                     \mathrm{Cat}\left( x^i_{s}; \frac{1-\alpha_{s}}{1-\alpha_t}\delta(\cdot , m) + \frac{\alpha_{s} - \alpha_t}{1-\alpha_t}\mu^\theta(x_t)[i, \cdot]\right) & \text{if } x^i_t = m    \\
                                     \mathrm{Cat}(x^i_s; \delta(\cdot, x^i_t))                                                                                                               & \text{if } x^i_t \neq m
                                 \end{cases}
\end{align}

Let $\mathrm{CE}(i, p)$ denote the cross entropy loss with sample $i$ and probability $p$: $\mathrm{CE}(i,p) = -\log p[i]$. The goal is to train $\mu^\theta(x_t)$ to match the factored posterior $\prod_{i=1}^L p(x_0^i \mid x_t)$. For such a denoiser, the transitions in \cref{eq:reverse_transition} correctly reverse the noising process and recover the data distribution at time $0$ \citep{sahoo2024simple,shi2024simplified}. The training objective for $\mu^\theta$ is the weighted cross entropy loss:
\begin{align}
    \label{eq:mdm_obj}
    \mathcal{L}_{\mathrm{MDM}}(\theta) & = \mathbb{E}_{t \sim \mathcal{U}[0,1], x_t \sim p(x_t | x_0), x_0 \sim p_{\mathrm{data}}}\left[\frac{-\alpha'_t}{1-\alpha_t}\sum^L_{i=1} \delta(x^i_t, m) \mathrm{CE}(x^i_{0}, \mu^\theta(x_t)[i, \cdot]) \right]
\end{align}
With a trained denoiser, the generation process consists of starting with a completely masked sequence $x_1 = M$ and iteratively unmasking tokens in the sequence over $T$ steps to obtain a final sample $x_0$. One step of this sampling procedure is done by simulating \cref{eq:reverse_transition}, which involves:
\begin{enumerate}[label=(\roman*)]
    \item Sampling an approximation of clean data from the denoiser $\hat{x}_0 \sim \mu^\theta(x_t)$
    \item Randomly choosing which (currently masked) positions $I$ in $x_t$ to unmask (by replacing $x^i_t = m$ with $\hat{x}^i_0$ for all $i \in I$). 
\end{enumerate}

Properly simulating \cref{eq:reverse_transition} requires random selection of the set of unmasking positions $I$ \citep{sahoo2024simple}. However, a number of works have found success in selecting $I$ through some heuristic informed by the denoiser logits $\mu^\theta(x_t)$ \citep{nie2025llada, kim2025trainworstplanbest,benhamu2025entropy_unmasking}. These involve calculating a score $s_i$ and then prioritizing unmasking positions $i$ with higher score (possibly with added noise).

Some options for the score calculation which have been investigated in previous work include choices such as (i) \textbf{High confidence}: The probability of the sampled clean tokens $s_i = \mu^\theta(x_t)[i, \hat{x}^i_0]$ \citep{nie2025llada}, (ii) \textbf{Margin}: The probability margin, i.e. the gap between the highest probability and the second highest probability $s_i = \mu^\theta(x_t)[i, j_1] - \mu^\theta(x_t)[i, j_2]$ where $j_1 = \arg\max_j \mu^\theta(x_t)[i, j]$ and $j_2 = \arg\max_{j \neq j_1} \mu^\theta(x_t)[i, j]$ \citep{kim2025trainworstplanbest}, and (iii) \textbf{Entropy}: The negative entropy of the position $s_i = -H(\mu^\theta(x_t)[i, \cdot])$ \citep{benhamu2025entropy_unmasking}.

These heuristics have been shown to yield better performance on a number of downstream tasks (such as coding and math), or on tasks such as Sudoku, where generating a valid solution is highly dependent on the order of unmasking \citep{nie2025llada, kim2025trainworstplanbest}.

The problem of selecting the unmasking ordering also informs the efficiency of performing inference, since being able to unmask multiple tokens leads to fewer calls to the denoising model to generate a complete sequence. Intuitively, we expect certain token positions to be independent of others, and we would expect unmasking them in parallel to retain the same performance as unmasking one at a time.

\subsection{Learnable Adaptive Order Policies}

Given the importance of the unmasking ordering, we ask the following question: armed with a dataset, and a pre-trained denoiser $\mu^\theta$, can we train a lightweight \emph{policy network} $q^\phi(i \mid x_t)$ which outputs a distribution over the next token position $i$ to unmask for $x_t$? The hope is to add a small number of additional trainable parameters, and spend some additional training iterations to obtain better performance on challenging tasks.

We can note that MDMs are sensitive to the token ordering precisely due to imperfections in the denoiser model. As argued by \citet{benhamu2025entropy_unmasking} a perfect denoiser is able to sample from the target regardless of the unmasking order (since it corresponds to different factorizations of the joint distribution, according to the chain rule of probability). Therefore, a reasonable objective for the policy is one that accounts for the loss of the denoiser model. Based on this, we propose to train the policy to place higher probability on positions which obtain a smaller loss, as measured by the cross entropy. This is captured in the objective:
\begin{align}
    \label{eq:order_obj}
    \mathcal{L}_{\mathrm{ORDER}}(\phi) & = \mathbb{E}_{t \sim \mathcal{U}[0,1], x_t \sim p(x_t | x_0), x_0 \sim p_{\mathrm{data}}}\left[\frac{-\alpha'_t}{1-\alpha_t} \sum^L_{i=1} \blue{q^\phi(i \mid x_t)}\mathrm{CE}(x^i_{0}, \mu^\theta(x_t)[i, \cdot]) \right]
\end{align}
Note that we assume $q(i \mid x_t) = 0$ for unmasked positions $x^i_t \neq m$. This is identical to the MDM objective \cref{eq:mdm_obj} except for the fact that the sum over masked positions is weighted by the policy $q^\phi$. In particular, for a uniform policy over masked tokens, we recover, up to multiplication, the vanilla MDM loss. 

We can further justify this choice of objective by relating it to an ELBO bound for the policy-guided unmasking process. Using an ELBO decomposition for such a process, derived by \citet{peng2025plannerawarepathlearning}, we can show that our loss in \cref{eq:order_obj} results from making approximations to avoid computing intractable terms. We present the details of this derivation in \cref{app:elbo_relation}. 

For a frozen denoiser network $\theta$, the policy network $q^\phi$ is trained to predict where the current denoiser is most likely correct. The optimal policy places all probability on the position with smallest cross entropy loss. We will refer to this theoretically optimal policy as the \textbf{oracle}: $q^\mathrm{oracle}(i \mid x_t) = \delta(i, \arg\min_{j, x^j_t = m} \mathrm{CE}(x^j_{0}, \mu^\theta(x_t)[j, \cdot]))$. We empirically validate the choice of this loss by evaluating the oracle policy (with access to ground truth data $x_0$), and confirming that it improves metrics relative to other heuristic samplers in \cref{sec:experiment}. Other approaches for training unmasking policies typically rely on more expensive gradient estimation, or RL procedures for tasks involving a reward. These are discussed in \cref{app:related_works}.

Finally we note that for sampling multiple positions, we can use the policy probabilities as score values $s_i = q^\phi(i \mid x_t)$ and use them in a similar way to other heuristics (for instance, unmasking the $k$ positions with the largest policy probabilities).

\subsection{Policy-Aware Denoiser Training}
\label{sec:joint_train_desc}

The policy objective in \cref{eq:order_obj} answers where the model should unmask next, but it also raises two related questions: can the policy improve the denoiser itself, and can the denoiser be trained while explicitly knowing that a policy will be used at sampling time? We note that the objective in \cref{eq:order_obj} already allows for gradients with respect to the denoiser parameters $\theta$, and admits a sensible interpretation. Namely, viewed from the perspective of training the denoiser, the objective upweights the loss at positions proportional to the probability the policy will select them for unmasking. 

A slight issue may occur with directly using \cref{eq:order_obj} for training the denoiser, since the policy may quickly collapse to a handful of positions and prevent denoiser training on most masked tokens. We alleviate this with a simple modification: we add the original MDM loss to the policy-weighted objective to ensure better training signal. This is a stability trick used in other works \citep{peng2025plannerawarepathlearning}.

Putting together these observations, we propose the following denoiser loss as a substitute to the vanilla MDM objective:
\begin{align}
    \label{eq:policy_aware_mdm_obj}
    \mathcal{L}_{\mathrm{PA\mbox{-}MDM}}(\theta)
     & =
    \mathbb{E}\left[
        \frac{-\alpha'_t}{1-\alpha_t}
        \sum_{i=1}^{L}
        \delta(x_t^i,m)
        \left(1 + q^\phi(i \mid x_t)\right)
        \mathrm{CE}(x_0^i,\mu^\theta(x_t)[i,\cdot])
        \right].
\end{align}
In implementation, the policy probabilities are detached inside the denoiser loss, so gradients with respect to $\theta$ do not backpropagate through $q^\phi$.

The policy itself is still trained with \cref{eq:order_obj}, with the denoiser losses detached when optimizing $\phi$. Joint training therefore lets the policy identify consequential positions, while simultaneously teaching the denoiser to allocate more capacity to them. The algorithm for either policy-only training or joint training is summarized in \cref{alg:training_policy_joint}.

To separate the effect of \emph{learned} policy guidance from the effect of simply reweighting the denoiser loss, we can also compare against heuristic-aware denoiser objectives, similar to those in \citep{peng2025plannerawarepathlearning}. Concretely, we can replace $q^\phi(i \mid x_t)$ in \cref{eq:policy_aware_mdm_obj} by normalized weights derived from standard confidence scores computed from the denoiser logits, such as the confidence (log-probability of the sampled token) or the margin. This yields a matched control where the denoiser is still trained with emphasis on selected positions, but the emphasis is determined by a fixed heuristic rather than a learned policy.

Such a comparison isolates the main question of interest: whether informing the denoiser about the existence of a learned unmasking policy during training provides benefits beyond the gains obtainable from generic heuristic reweighting alone.

\begin{algorithm}[t]
\small
\caption{Training with Policy-Only or Joint Updates}
\label{alg:training_policy_joint}
\DontPrintSemicolon
\SetKwComment{tcp}{// }{}
\KwIn{dataset $\mathcal{D}$, denoiser $\mu^\theta$, policy $q^\phi$, noise schedule $\alpha_t$}

\While{not converged}{
    Sample $x_0 \sim \mathcal{D}$, $t \sim \mathcal{U}[0,1]$, and $x_t \sim p(x_t \mid x_0)$\;
    Compute denoiser logits $\mu^\theta(x_t)$ and policy probabilities $q^\phi(i \mid x_t)$ over masked positions\;
    Compute token losses $\ell_i \gets \delta(x_t^i,m)\,\mathrm{CE}(x_0^i,\mu^\theta(x_t)[i,\cdot])$ for $i=1,\dots,L$\;
    Define
    \[
        \mathcal{L}_{\mathrm{ORDER}}(\phi)
        =
        \frac{-\alpha_t'}{1-\alpha_t}
        \sum_{i=1}^{L} q^\phi(i \mid x_t)\,\operatorname{stopgrad}(\ell_i)
    \]
    \uIf{joint training is enabled}{
        Define
        \[
            \mathcal{L}_{\mathrm{PA\mbox{-}MDM}}(\theta)
            =
            \frac{-\alpha_t'}{1-\alpha_t}
            \sum_{i=1}^{L} \left(1 + \operatorname{stopgrad}(q^\phi(i \mid x_t))\right)\ell_i
        \]
        Define $\mathcal{L}_{\mathrm{total}} = \mathcal{L}_{\mathrm{ORDER}}(\phi) + \mathcal{L}_{\mathrm{PA\mbox{-}MDM}}(\theta)$\tcp*[r]{joint training}
    }
    \Else{
        Define $\mathcal{L}_{\mathrm{total}} = \mathcal{L}_{\mathrm{ORDER}}(\phi)$\tcp*[r]{policy-only training}
    }
    Update trainable parameters using $\mathcal{L}_{\mathrm{total}}$\;
}
\end{algorithm}

\section{Experiments}
\label{sec:experiment}

We evaluate our adaptive ordering policy approach on two constraint-satisfaction tasks known to be sensitive to token ordering, Sudoku puzzle solving and 3-SAT (Boolean satisfiability with 3 literals per clause) \citep{kim2025trainworstplanbest, ye2024beyond}; protein sequence generation with DPLM \citep{wang2024dplm}; code generation with Open-dLLM-0.5B \citep{opendllm2025}; and six language and reasoning benchmarks with LLaDA-8B \citep{nie2025llada}.

\paragraph{Policy architecture} We parameterize $q^\phi$ as a lightweight per-token MLP that conditions on both the denoiser's confidence scores (max log-probability) and the hidden states from the last layer of the base model. Specifically, each scalar confidence score is projected to the hidden dimension ($d{=}384$) via a linear layer, summed with the corresponding hidden-state vector, and passed through a two-layer MLP (hidden dimension 128, ReLU activation) that outputs a per-position routing logit. The policy adds ${\sim}$50K parameters ({$<$}1\% of the base model). In the policy-only setting, it is trained on top of a frozen denoiser using \cref{eq:order_obj}, while in the joint-training setting it is optimized together with the denoiser under the objectives described in \cref{sec:joint_train_desc}. We also experimented with a transformer-based policy variant \citep{jazbec2025learningunmasking}, which did not yield further improvement (see \cref{app:transformer_policy}).

\paragraph{Baselines} At inference, we compare against baselines such as \textbf{High conf.}, \textbf{Margin}, \textbf{Entropy}, \textbf{Random}, and \textbf{Oracle} ordering strategies described in \cref{sec:method}, based on applicability and relevance. For our joint training experiments, we train using the modified denoiser objective from \cref{eq:policy_aware_mdm_obj}, with baselines being trained using heuristic weightings, as described in \cref{sec:joint_train_desc}.


\paragraph{Task Details} For Sudoku and 3-SAT we use a 6M-parameter GPT-2 denoiser trained as an MDM with $20$ and $10$ denoising steps respectively. For protein evaluation we adapt the DPLM-150M model \citep{wang2024diffusionlanguagemodelsversatile} and evaluate it for both policy-only and joint policy-denoiser training. For large-scale settings, we do policy-only training with Open-dLLM-0.5B for code generation, and LLaDA-8B across code, reasoning, question answering, translation, and summarization tasks. The Open-dLLM policy is trained on FineCode, while the LLaDA policy uses a mixture of general web, mathematical, and code data. Full training and evaluation details are provided in \cref{app:experimental_details}.

\subsection{Results on Structured Data}

\begin{wraptable}{r}{0.58\textwidth}
    \vspace{-0.8em}
    \caption{Main results for policy-only and joint training. We report mean accuracy for Sudoku and 3-SAT, and mean pLDDT on sequences of length $\{100, 200, 300, 400, 500\}$ for DPLM-150M. All results are averaged over 3 seeds.}
    \label{tab:main_results_combined}
    \centering
    \small
    \renewcommand{\arraystretch}{1.06}
    \resizebox{\linewidth}{!}{%
    \begin{tabular}{lccc}
        \toprule
        & \textbf{Sudoku} & \textbf{3-SAT} & \textbf{DPLM-150M} \\
        \midrule
        \multicolumn{4}{l}{\textbf{Policy-only training}} \\
        \midrule
        Random & 07.33 $\pm$ 0.07 & 87.23 $\pm$ 0.74 & 40.04 $\pm$ 0.78 \\
        Entropy & 25.48 $\pm$ 0.15 & 87.63 $\pm$ 0.25 & 55.88 $\pm$ 0.93 \\
        High conf.                    & 18.71 $\pm$ 0.16 & 86.87 $\pm$ 0.76 & 82.20 $\pm$ 0.76 \\
        Margin                        & 91.53 $\pm$ 0.04 & 88.33 $\pm$ 0.15 & 82.02 $\pm$ 0.31 \\
        Policy                        & \textbf{93.75 $\pm$ 0.20} & \textbf{89.00 $\pm$ 0.00} & \textbf{83.85 $\pm$ 0.80} \\ \noalign{\vspace{2pt}}\hdashline\noalign{\vspace{2pt}}
        Oracle policy                 & 100.00 $\pm$ 0.00 & 93.90 $\pm$ 0.00 & N/A \\
        \midrule
        \multicolumn{4}{l}{\textbf{Joint training}} \\
        \midrule
        Base-model                   & 18.71 $\pm$ 0.16 & 86.87 $\pm$ 0.76 & 82.20 $\pm$ 0.76 \\
        High conf.                  & 19.44 $\pm$ 0.03 & 88.80 $\pm$ 0.26 & 82.85 $\pm$ 0.85 \\
        Margin                      & 90.23 $\pm$ 0.03 & 88.67 $\pm$ 0.15 & 82.28 $\pm$ 0.53 \\
        Policy                      & \textbf{93.52 $\pm$ 0.11} & \textbf{90.57 $\pm$ 0.06} & \textbf{84.94 $\pm$ 1.00} \\
        \bottomrule
    \end{tabular}}

    \vspace{0.2em}

    {\footnotesize The policy adds {$<$}50K parameters ({$<$}1\% of the base model).}
    \vspace{-0.6em}
\end{wraptable}

\begin{figure}[t]
    \centering
    \includegraphics[width=\linewidth]{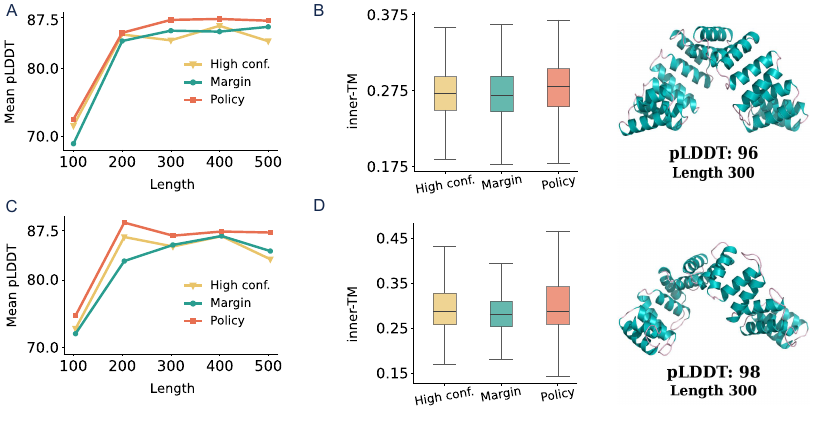}
    \caption{Adaptive ordering on DPLM-150M. Top (A, B): policy-only adaptation. Bottom (C, D): joint training. Left: mean pLDDT by target sequence length. Right: inner-TM diversity, where lower is better. Representative predicted 3D folds illustrate jointly trained samples across different sequence lengths. The learned policy improves foldability while maintaining diversity close to the heuristic baselines.}
    \label{fig:dplm_policy_only}
\end{figure}

\noindent\textbf{Policy-only training.} The upper block of \Cref{tab:main_results_combined} summarizes policy-only adaptation with a frozen denoiser using \cref{eq:order_obj}. Across Sudoku, 3-SAT, and protein generation with DPLM, the learned policy outperforms the heuristic baselines. On DPLM-150M, the policy improves predicted foldability while maintaining diversity close to the confidence and margin orderings, as also shown in the top row of \cref{fig:dplm_policy_only}. This establishes that even without modifying the base denoiser, learning the unmasking order alone yields consistent gains across domains.

\noindent\textbf{Joint training.} The lower block of \Cref{tab:main_results_combined} evaluates the policy-aware denoiser objective from \cref{eq:policy_aware_mdm_obj}. Here, \emph{High conf.} and \emph{Margin} replace the learned policy weights with normalized weights derived from the confidence and margin heuristics respectively. The main pattern is that learned policy reweighting improves the denoiser more reliably than matched heuristic reweighting. On Sudoku and 3-SAT, policy-aware scaling gives the strongest stochastic results among the compared training objectives. The same comparison also extends to DPLM, where the learned policy-weighted objective improves protein generation relative to the heuristic-weighted controls, as also illustrated in the bottom row of \Cref{fig:dplm_policy_only}. Taken together, these results show that the policy is useful not only as an inference-time ordering policy but also as a training signal for the denoiser.

\paragraph{Protein sequence generation} For protein generation, we conduct a more comprehensive evaluation of our joint policy-denoiser model by generating 100 sequences at lengths 200, 300, \ldots, 800, which are then folded into 3D structures using ESMFold \citep{lin2022evolutionary}. Results are shown in \Cref{tab:papl_protein_compare}, which reports structure quality, foldability, and diversity metrics (the precise metric definitions collected in \cref{app:protein_metrics}). We note that both our results and those of \citet{peng2025plannerawarepathlearning} reported in the table use the same number of parameters (150M).

\begin{table}[H]
    \caption{Protein sequence generation results including our joint-policy, with baselines results taken from \citet{peng2025plannerawarepathlearning}. Each model generates 100 sequences at lengths 200, 300, \ldots, 800, which are folded into 3D structures using ESMFold. Structure quality is measured by pLDDT, pTM, and pAE, while diversity is measured by token entropy and sequence uniqueness. Foldability is the percent of sequences with pLDDT $> 80$, pTM $> 0.7$, and pAE $< 10$.}
    \label{tab:papl_protein_compare}
    \centering
    \scriptsize
    \resizebox{0.95\textwidth}{!}{
    \begin{tabular}{lcccccc}
        \toprule
        \textbf{Model} & \textbf{pLDDT} $\uparrow$ & \textbf{pTM} $\uparrow$ & \textbf{pAE} $\downarrow$ & \textbf{Foldability (\%)} $\uparrow$ & \textbf{Entropy} $\uparrow$ & \textbf{Diversity (\%)} $\uparrow$ \\
        \midrule
        \multicolumn{7}{l}{\textbf{Large}} \\
        ESM3 & 34.13 & 0.23 & 24.65 & 1.50 & 3.99 & \textbf{93.44} \\
        ProGen2-medium & 57.94 & 0.38 & 20.81 & 12.75 & 2.91 & 91.45 \\
        ProGen2-large & 55.07 & 0.35 & 22.00 & 11.87 & 2.73 & 91.48 \\
        DPLM-650M & 79.53 & 0.66 & 11.85 & 49.14 & 3.18 & 92.22 \\
        \midrule
        \multicolumn{7}{l}{\textbf{150M-scale}} \\
        EvoDiff & 31.84 & 0.21 & 24.76 & 0.43 & 4.05 & 93.19 \\
        ProGen2-small & 49.38 & 0.28 & 23.38 & 4.48 & 2.55 & 89.31 \\
        DPLM-150M & 80.23 & 0.65 & 12.07 & 48.14 & 3.14 & 92.80 \\
        DLM-150M & 81.32 & 0.65 & 12.00 & 42.43 & 3.21 & 92.45 \\
        DLM-150M + PAPL & 81.48 & 0.72 & \textbf{8.97} & \textbf{59.40} & 3.12 & 91.73 \\
        Joint policy (ours) & \textbf{86.43} & \textbf{0.76} & 9.68 & 54.14 & \textbf{4.12} & 93.06 \\
        \bottomrule
    \end{tabular}
    }
\end{table}

\Needspace{0.30\textheight}
\begin{wrapfigure}{r}{0.54\textwidth}
    \vspace{-1.1em}
    \centering
    \includegraphics[width=\linewidth]{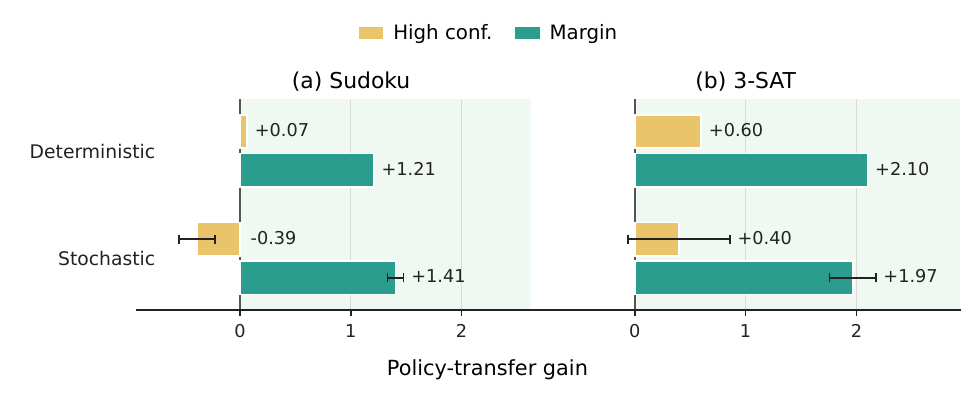}
    \captionsetup{skip=2pt}
    \caption{Transfer from policy-weighted training to heuristic decoding. We plot the accuracy difference between the denoiser trained with policy-weighted and heuristic weighted joint objectives, and when decoding with the matched heuristic. Policy-weighted training improves both heuristics.}
    \label{fig:heuristic_transfer}
    \vspace{-0.7em}
\end{wrapfigure}
\noindent\textbf{Transfer to heuristic decoding.} An important property of policy-aware training is that its benefits are not limited to decoding with the learned policy itself. As shown in \cref{fig:heuristic_transfer}, the denoiser trained with policy-aware scaling also improves heuristic decoding on both tasks. This suggests that the policy-weighted objective does not merely tailor the denoiser to one decoding rule, but improves the underlying denoiser in a way that transfers across different unmasking heuristics.
\WFclear

\subsection{Ablations}
\begin{wrapfigure}{r}{0.59\textwidth}
    \vspace{-0.8em}
    \centering
    \includegraphics[width=\linewidth]{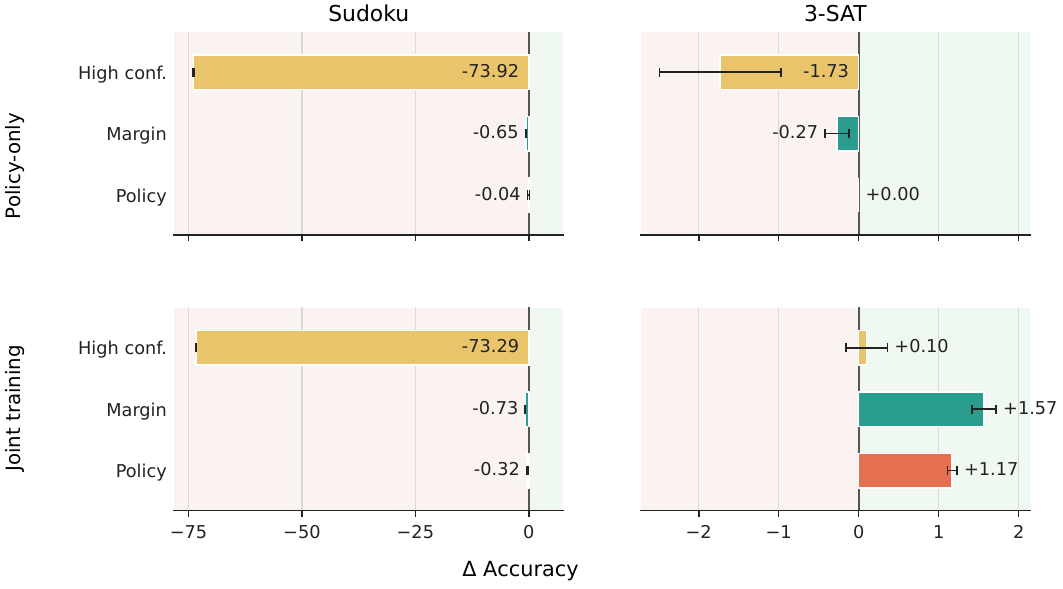}
    \captionsetup{skip=2pt}
    \caption{Deterministic and stochastic decoding on combinatorial tasks. We compute $\Delta$ as the difference between stochastic accuracy and deterministic accuracy. Negative values indicate degradation under stochastic decoding.}
    \label{fig:decoding_sensitivity}
    \vspace{-0.7em}
\end{wrapfigure}

\noindent\textbf{Deterministic versus stochastic decoding.} We study the effect of decoding noise in \Cref{fig:decoding_sensitivity}. We evaluate the methods on different \emph{decoding strategies}: namely deterministic decoding which unmasks the highest scoring position each step, and stochastic, which adds Gumbel noise to this. One notable observation is the discrepancy between high-confidence Sudoku accuracy under deterministic and stochastic decoding. The latter is reported by \citet{kim2025trainworstplanbest}, who propose the margin heuristic as an improvement. We find that this gap is primarily due to the sensitivity of the high-confidence heuristic to the decoding strategy. In fact, with deterministic decoding, it outperforms the margin heuristic. The margin heuristic is less sensitive to decoding. Our learned policy is similarly robust across decoding regimes. The lower block shows the same pattern in joint training. Therefore, the policy-weighted objective avoids pathological changes in performance due to decoding parameters, while outperforming the alternative methods.

\begingroup
\setlength{\intextsep}{6pt}
\begin{figure}[t]
    \centering
    \captionsetup{skip=3pt}
    \includegraphics[width=0.9\linewidth]{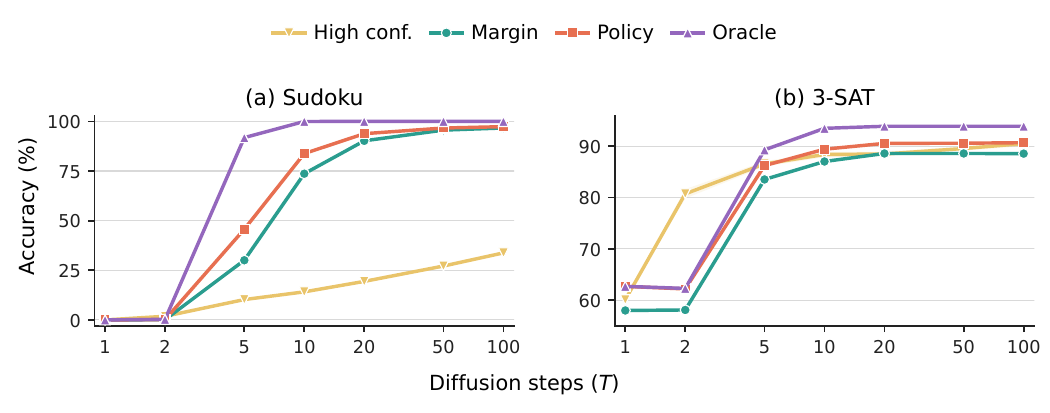}
    \caption{Accuracy as a function of the number of reverse diffusion steps $T$ on Sudoku (left) and 3-SAT (right). Better ordering is especially valuable at small step budgets, where improved unmasking policies can recover substantially more accuracy for the same inference cost.}
    \label{fig:sweep_t_efficiency}
\end{figure}
\endgroup

\paragraph{Efficiency as a function of diffusion steps} We also study how ordering quality interacts with the number of reverse steps $T$, which directly controls inference cost. As shown in \cref{fig:sweep_t_efficiency}, the learned ordering improves efficiency by achieving higher accuracy for the same step budget.
On Sudoku, the learned policy substantially outperforms both heuristic baselines and nearly matches the oracle at $T{=}100$. On 3-SAT, the learned policy consistently improves over the high-confidence heuristic and is competitive with, or slightly better than, the margin heuristic across moderate and large step budgets. In both tasks, the oracle remains better than the learned policy, indicating that there is still substantial room to improve learned unmasking strategies. Overall, these results show that ordering affects both accuracy and inference efficiency. 

\begin{wraptable}{r}{0.45\textwidth}
    \vspace{-0.8em}
    \caption{Comparison with continued training on the denoiser.}
    \label{tab:additional_train_policy_baselines}
    \centering
    \small
    \renewcommand{\arraystretch}{1.06}
    \resizebox{\linewidth}{!}{%
    \begin{tabular}{lc}
        \toprule
        \textbf{Ordering} & \textbf{Sudoku} \\
        \midrule
        High Conf. & 18.71 $\pm$ 0.16 \\
        Margin & 91.53 $\pm$ 0.04  \\
        High Conf. (Continued) & 18.73 $\pm$ 0.14 \\
        Margin (Continued) & 91.52 $\pm$ 0.04 \\
        Policy & \textbf{93.75 $\pm$ 0.20} \\
        \bottomrule
    \end{tabular}}
    \vspace{-0.6em}
\end{wraptable}

\noindent\textbf{Comparison with additional denoiser training} 
 Given a pre-trained denoiser network, one could choose to improve the model either by training the policy-network and freezing the denoiser (as in our algorithm), or simply continuing to train the denoiser (with the standard MDM loss). We compare  against this alternative for the Sudoku task, with a matched training budget, in \cref{tab:additional_train_policy_baselines}. We observe that continued denoiser training yields little improvement, with both the high-confidence and margin heuristics. In contrast, the policy-only training more substantially improves the accuracy. We also note that updating the lightweight policy is cheaper per step than full denoiser training (owing to fewer parameters).

\subsection{Language and code generation}

\begin{table}[H]
    \caption{Best pass@1 (\%) on Open-dLLM-0.5B code benchmarks.}
    \label{tab:open_dllm}
    \centering
    \small
    \renewcommand{\arraystretch}{1.06}
    \begin{tabular*}{\linewidth}{@{\extracolsep{\fill}}lcccc@{}}
        \toprule
        \textbf{Ordering} & \textbf{HumanEval} & \textbf{HumanEval+} & \textbf{MBPP} & \textbf{MBPP+} \\
        \midrule
        High conf. & 17.62 $\pm$ 2.45 & 15.61 $\pm$ 2.34 & 15.58 $\pm$ 1.24 & \textbf{23.28 $\pm$ 1.65} \\
        Margin & 12.56 $\pm$ 1.78 & 11.65 $\pm$ 1.79 & 8.72 $\pm$ 0.70 & 13.47 $\pm$ 1.00 \\
        Policy & \textbf{18.90 $\pm$ 3.00} & \textbf{16.46 $\pm$ 2.90} & \textbf{16.20 $\pm$ 1.31} & 22.80 $\pm$ 1.76 \\
        \bottomrule
    \end{tabular*}
\end{table}

\noindent\textbf{Open-dLLM code generation.} We train only the policy on the 0.5B-parameter Open-dLLM code model, whose Qwen2-based backbone has a vocabulary of 151,936 tokens \citep{opendllm2025,qwen2}. \Cref{tab:open_dllm} reports the best pass@1 obtained for each ordering method on each benchmark after independently sweeping its decoding hyperparameters. The policy outperforms high-confidence ordering on three of four datasets and margin ordering on all four.

\begin{wraptable}{r}{0.66\textwidth}
    \vspace{-1.1em}
    \caption{Policy-only training results on LLaDA-8B. The top table decodes with 1 token/step, while the bottom one does 4 tokens/step}
    \label{tab:llada_results}
    \centering
    \footnotesize
    \setlength{\tabcolsep}{2.2pt}
    \renewcommand{\arraystretch}{1.03}
    \begin{adjustbox}{max width=\linewidth}
    \begin{tabular}{lcccccc}
        \toprule
        \textbf{Ordering} & \textbf{HumanEval} & \textbf{BBH} & \textbf{CoQA} & \textbf{WMT16} & \textbf{MBPP} & \textbf{XSum} \\
        \midrule
        High conf. & \textbf{34.15} & \textbf{44.82} & \textbf{80.48} & 12.031 & 39.60 & 23.051 \\
        Policy & \textbf{34.15} & 43.96 & 78.83 & \textbf{12.076} & \textbf{39.80} & \textbf{23.265} \\
        \bottomrule
    \end{tabular}
    \end{adjustbox}

    \vspace{0.35em}

    \begin{adjustbox}{max width=\linewidth}
    \begin{tabular}{lcccccc}
        \toprule
        \textbf{Ordering} & \textbf{HumanEval} & \textbf{BBH} & \textbf{CoQA} & \textbf{WMT16} & \textbf{MBPP} & \textbf{XSum} \\
        \midrule
        High conf. & 12.80 & \textbf{41.94} & 76.93 & 7.666 & \textbf{16.00} & 22.380 \\
        Policy & \textbf{14.02} & 41.41 & \textbf{77.34} & \textbf{8.020} & 14.60 & \textbf{22.479} \\
        \bottomrule
    \end{tabular}
    \end{adjustbox}
    \vspace{-0.6em}
\end{wraptable}

\noindent\textbf{LLaDA language and reasoning.} We additionally train the policy on LLaDA-8B and evaluate on various coding and language tasks (with details on tasks and metrics in \cref{app:language_experiments}). The policy outperforms high-confidence ordering on three and ties on one (\cref{tab:llada_results}). We further evaluate the policy using $1/4$th the denoising steps. The policy's relative performance improves on four of the six tasks, reinforcing the earlier observation that the learned policy supports few-step inference better than competing heuristics. Importantly, our supervised objective uses a general (non-task specific) training set, and does not require constructing a task-specific reward. This renders the method applicable to code and math tasks, where verifiable rewards may be available, as well as question answering, translation, and summarization, where reward design is less direct.

\subsection{Discussion}

Several observations emerge from these results. First, the learned policy improves over heuristic orderings on both structured tasks (constraint-satisfaction, proteins) and on broader text tasks (coding, language), using a lightweight auxiliary network with less than 1\% parameter overhead. The latter setting demosntrates the scalability of our method to large model and vocabulary sizes. Second, our experiments suggest that the learned policy is more amenable to parallel decoding, and generally outperforms other unmasking strategies at fewer diffusion steps. Third, the policy is useful both as an inference-time ordering rule and as a training signal for the denoiser (even when using other ordering strategies).
Finally, the remaining gap to the oracle suggests that better policies are still achievable.  
\section{Related Works}
\label{app:related_works}

\citet{wang2025learningorderar} also train a policy for the unmasking order, by treating the order as a latent variable $z$. They propose a variational method for optimizing it, which requires parameterizing the posterior approximation $q^\phi(z \mid x)$, in addition to the trainable policy over orders $p^\theta(z\mid x)$. The former is only used as part of the training objective for the latter, and not during inference. In addition, the optimization of the variational posterior requires gradient estimation techniques to reduce variance, and complicates the optimization loop. Our loss by contrast is much simpler. 

Another set of works assume access to verifiable reward functions rather than a dataset \citep{hong2025improvingdiscretediffusionunmasking, jazbec2025learningunmasking}. These frame the generation process of a masked diffusion model as a Markov Decision Process and optimize the unmasking policy using RL objectives. By contrast, our supervised objective requires data but no task-specific reward construction. It therefore supports both domains where verifiable rewards could be defined, such as code and mathematical reasoning, and domains where a suitable reward is less direct such as summarization and protein generation.

\citet{peng2025plannerawarepathlearning} propose a modification of the MDM loss which accounts for using heuristics when determining the unmasking order (rather than random unmasking). The loss resembles our objective \cref{eq:policy_aware_mdm_obj}, and we outline the connections between our objective and their ELBO bound in \cref{app:elbo_relation}. The main difference with our work is that our objective makes necessary simplifications to enable joint training of the denoiser and policy (as opposed to only training the denoiser, as done in their work). 

Recent works also learn auxiliary token-level scoring modules that can be used to remask tokens during inference, to improve generation quality. \citet{huang2025dontsettle} train a confidence head for self-reflective remasking in diffusion language models: for unmasked tokens in $x_t$, the head predicts whether they are correct, while for masked positions it is supervised by the denoiser's probability of recovering the ground-truth token. \citet{meshchaninov2025guided} similarly fine-tune a lightweight error-prediction layer on top of a pretrained model,  to predict whether tokens in a denoiser-produced $\hat{x}_0$ are incorrect. These works are related to our policy-only setting in that they train an additional scoring layer on top of a pretrained denoiser. However, while their framework is concerned with using scores to remask tokens, our method more directly aims to learn an unmasking order. Our work differs in the objective and training setup: we directly optimize a policy over unmasking decisions with a cross-entropy-based objective weighted by denoiser losses, rather than training an error classifier, and we additionally study a joint training objective that couples the policy and denoiser, which neither of these approaches explore.

\section{Conclusion}
\label{sec:conclusion}

This work studies a method for training a policy over token orderings for masked diffusion models, by learning to sample positions with lower cross-entropy loss. We demonstrate that our approach outperforms common heuristics on logical tasks that are sensitive to the unmasking order, and that the same adaptation transfers to protein generation with DPLM as well as code, language, and reasoning benchmarks with Open-dLLM and LLaDA. In the policy-only setting, these gains require only a lightweight auxiliary network, while joint training further improves performance through a policy-aware denoising objective. Our results suggest that the unmasking order in masked diffusion models should be treated as a learnable component rather than a fixed heuristic. In addition to improved task performance, the proposed method is observed to yield benefits such as improving efficiency under limited diffusion-step budgets.

Future work could explore possible training-free, inference-time extensions (alleviating the need for a dataset) as well as alternate policy parameterizations and training schemes to close the gap between current and oracle performance. 

\newpage

\subsection*{AI use statement}

In this work, we used generative AI tools for editing the writing (eg. for polishing grammar, word-choice), as well as assistance in generating and editing code implementations and plotting scripts. The authors reviewed all AI-assisted text,
inspected and ran the generated code, and checked its outputs against the intended behavior. 

We have not used generative AI tools for helping develop theoretical models or conceptual frameworks, formulating or proving mathematical claims, refining hypotheses, providing feedback on research methodology or to interpret experimental results,
and others research subtasks such as generating synthetic data, are not applicable to this work.

We take responsibility for the final content of this work, including all text, claims, code, and artifacts produced with the aid of generative AI.
\subsection*{Ethics statement}
\label{sec:ethics}

The work proposes a method for improving masked diffusion models by learning the unmasking order. This is a general algorithm which has potential uses in biological applications (e.g. generating protein or DNA sequences) as well as potential harmful consequences such as for instance, in generating code with exploitable weaknesses. However, we discourage such applications.

\subsection*{Reproducibility statement}




We include experimental details in the main paper and in \cref{app:experimental_details} (including compute resources, metrics, training schemes etc.). All data and models trained on or fine-tuned are publicly available.

\subsubsection*{Acknowledgements}
\label{sec:acknowledgements}
\looseness=-1

The research was enabled in part by computational resources provided by the Digital Research Alliance of Canada (\url{https://alliancecan.ca}) and Mila (\url{https://mila.quebec}).

\bibliography{references}
\bibliographystyle{iclr2027_conference}

\newpage
\appendix
\section{Experimental Details}
\label[appendix]{app:experimental_details}

\subsection{Combinatorial Tasks}

\paragraph{Tasks and datasets} We consider two tasks. \textbf{Sudoku}: 9${\times}$9 puzzles where the model fills in blank cells given a partially completed grid, represented as a flat sequence of 81 tokens (digits 1--9). \textbf{3-SAT}: Boolean satisfiability instances with 9 variables and 3 literals per clause, where the model must find a satisfying assignment given the clause structure. For Sudoku, we follow the dataset and evaluation protocol of \citet{kim2025trainworstplanbest}; for 3-SAT, we use the dataset of \citet{ye2024beyond}. We report instance-level accuracy: an instance is correct only if the entire solution is valid.

\paragraph{Base model and training} We use a 6M-parameter GPT-2 architecture (3 layers, 384 hidden dimensions, 12 attention heads, vocabulary size 31) as the denoiser $\mu^\theta$, trained as a masked diffusion model (MDM) with $T{=}20$ diffusion steps. Training uses focal-loss-style token reweighting ($\alpha{=}0.25$, $\gamma{=}1$) and linear time reweighting, with a learning rate of $10^{-3}$, batch size of 1024, cosine learning rate schedule, and mixed-precision (fp16) on a single A100 GPU. We study two training regimes on top of this base model. In the \emph{policy-only} regime, the denoiser is frozen and only the lightweight policy head is optimized using \cref{eq:order_obj}; for Sudoku, the policy is trained for 24K steps after 115K denoiser pretraining steps, and for 3-SAT, the policy is trained for 7.5K steps after 58.5K denoiser pretraining steps. In the \emph{joint-training} regime, we optimize the denoiser and policy together with the policy-aware weighted objective from \cref{eq:policy_aware_mdm_obj}.

\noindent\textbf{Deterministic and stochastic decoding.} \Cref{tab:det_stoch} includes the accuracies for both deterministic and stochastic decoding (which were used to compute differences for \cref{fig:decoding_sensitivity}). Without adding noise the heuristics perform similarly and learned policy achieves best results on both decoding settings.
\begin{table}[H]
    \centering
    \captionsetup{width=0.82\textwidth}
    \caption{Deterministic and stochastic decoding accuracy (\%) on combinatorial tasks.}
    \label{tab:det_stoch}
    \footnotesize
    \renewcommand{\arraystretch}{1.0}
    \begin{tabular*}{0.82\textwidth}{@{\extracolsep{\fill}}lcccc@{}}
        \toprule
        & \multicolumn{2}{c}{\textbf{Sudoku}} & \multicolumn{2}{c}{\textbf{3-SAT}} \\
        \cmidrule(lr){2-3} \cmidrule(lr){4-5}
        \textbf{Ordering} & \textbf{Det.} & \textbf{Stoch.} & \textbf{Det.} & \textbf{Stoch.} \\
        \midrule
        \multicolumn{5}{l}{\textbf{Policy-only}} \\
        High conf. & 92.63 & 18.71 $\pm$ 0.16 & 88.6 & 86.87 $\pm$ 0.76 \\
        Margin & 92.18 & 91.53 $\pm$ 0.04 & 88.6 & 88.33 $\pm$ 0.15 \\
        Policy & \textbf{93.79} & \textbf{93.75} $\pm$ \textbf{0.20} & \textbf{89.0} & \textbf{89.00} $\pm$ \textbf{0.00} \\
        \midrule
        \multicolumn{5}{l}{\textbf{Joint training}} \\
        Base model & 92.63 & 18.71 $\pm$ 0.16 & 88.6 & 86.87 $\pm$ 0.76 \\
        High conf. & 92.73 & 19.44 $\pm$ 0.03 & 88.7 & 88.80 $\pm$ 0.26 \\
        Margin & 90.96 & 90.23 $\pm$ 0.03 & 87.1 & 88.67 $\pm$ 0.15 \\
        Policy & \textbf{93.84} & \textbf{93.52} $\pm$ \textbf{0.11} & \textbf{89.4} & \textbf{90.57} $\pm$ \textbf{0.06} \\
        \bottomrule
    \end{tabular*}
    \vspace{-0.8em}
\end{table}

\begin{wraptable}{r}{0.54\textwidth}
    \vspace{-1.2em}
    \caption{Held-out NLL under deterministic decoding. 3-SAT uses 10 denoising steps; Sudoku uses 20.}
    \label{tab:duel}
    \centering
    \footnotesize
    \renewcommand{\arraystretch}{1.0}
    \begin{adjustbox}{max width=\linewidth}
    \begin{tabular}{lcc}
        \toprule
        \textbf{Ordering} & \textbf{3-SAT NLL} & \textbf{Sudoku NLL} \\
        \midrule
        \multicolumn{3}{l}{\textbf{Policy-only}} \\
        Entropy & 0.028561 & 0.003708 \\
        High conf. & 0.028587 & 0.003711 \\
        Margin & 0.028613 & 0.004185 \\
        Random & 0.030919 & 0.063625 \\
        Policy & \textbf{0.027888} & \textbf{0.003323} \\
        \midrule
        \multicolumn{3}{l}{\textbf{Joint training}} \\
        High conf. & 0.018856 & 0.003649 \\
        Margin & 0.025383 & 0.004445 \\
        Policy & \textbf{0.018210} & \textbf{0.003056} \\
        \bottomrule
    \end{tabular}
    \end{adjustbox}
    \vspace{-0.6em}
\end{wraptable}

\noindent\textbf{Exact likelihood.} For the deterministic combinatorial samplers, we compute held-out negative log-likelihood using DUEL \citep{turok2026duel}, as an additional evaluation metric. As shown in \cref{tab:duel}, the learned policy gives the lowest NLL in both training regimes and on both tasks, demonstrating that the policy training better captures the dataset distribution (in addition to having better accuracy). We do not apply DUEL to DPLM because its token sampling is stochastic and its ordering scores depend on the sampled tokens, whereas DUEL requires deterministic position selection.

\subsection{Protein / DPLM Experiments}

\subsubsection{DPLM Setup}
\paragraph{Evaluation setup} For protein experiments, we adapt the same policy parameterization to DPLM-150M \citep{wang2024dplm}. The policy again takes per-position confidence scores together with final-layer hidden states and predicts routing logits over masked positions. Our DPLM setup supports both policy-only adaptation, where the pretrained backbone is frozen and only the lightweight policy head is trained, and joint training of the backbone and policy through the policy-aware weighted objective from \cref{eq:policy_aware_mdm_obj}. In the joint DPLM, we initialize from the released DPLM-150M checkpoint of \citet{wang2024dplm}\footnote{\url{https://huggingface.co/airkingbd/dplm_150m}}, unfreeze the backbone, and fine-tune with the adaptive DPLM training configuration using a reduced learning rate ($4\times 10^{-5}$) on A100 and H100 GPUs. For the DPLM results summarized in \Cref{tab:main_results_combined} and visualized in \Cref{fig:dplm_policy_only}, we generate proteins at sequence lengths $\{100,200,300,400,500\}$; pLDDT is averaged across three random seeds after averaging within seed over sequence lengths, and we additionally report inner-TM diversity in the corresponding diversity plots. For the protein sequence generation comparison in \Cref{tab:papl_protein_compare}, we follow \cite{peng2025plannerawarepathlearning} and generate 100 sequences at lengths 200, 300, \ldots, 800 and report pLDDT, pTM, pAE, foldability, token entropy, and sequence diversity. The heuristic baselines are high-confidence decoding and margin decoding, matching the orderings used in the combinatorial experiments.

\subsubsection{Protein Sequence Generation Metrics}
\label[appendix]{app:protein_metrics}
For protein evaluation, we fold each generated sequence with ESMFold \citep{lin2022evolutionary} and use the predicted structures to assess quality, foldability, and diversity. For the joint comparison in \Cref{tab:papl_protein_compare}, we report three structural confidence metrics, a composite foldability score, and two diversity statistics. The per-sequence structural metrics are:
\begin{align}
\mathrm{pLDDT}(i) &= 100 \times \mathbb{E}_{j \in \mathcal{N}(i)}\left[1 - \frac{|d^{\mathrm{pred}}_{ij} - d^{\mathrm{true}}_{ij}|}{d^{\mathrm{true}}_{ij}}\right], \\
\mathrm{pTM} &= \max_{u} \frac{1}{L} \sum_{i=1}^{L} \frac{1}{1 + \left(d_{i,u(i)} / d_0(L)\right)^2}, \\
\mathrm{pAE}(i,j) &= \mathbb{E}\!\left[\left\|x^{\mathrm{pred}}_i - x^{\mathrm{true}}_j\right\|_2\right].
\end{align}
pLDDT measures local per-residue confidence, pTM measures global structural similarity and is adapted from the TM-score \citep{Zhang2004ScoringFF}, and pAE measures expected alignment error across residue pairs. We report pLDDT averaged over residues and pAE averaged over residue pairs; higher pLDDT and pTM are better, while lower pAE is better. We also report a binary \emph{foldability} rate, defined as the fraction of generated sequences satisfying $\mathrm{pLDDT} > 80$, $\mathrm{pTM} > 0.7$, and $\mathrm{pAE} < 10$ (as in \citet{peng2025plannerawarepathlearning}).

To assess diversity and possible mode collapse, we report token entropy, sequence diversity, and inner-TM. Token entropy and sequence diversity are defined as:
\begin{align}
H &= - \sum_{a \in \mathcal{A}} p(a)\log p(a), \\
\mathrm{Id}\!\left(x^{(m)}, x^{(n)}\right) &= \frac{1}{L}\sum_{i=1}^{L} \mathbf{1}\!\left[x^{(m)}_i = x^{(n)}_i\right], \\
\mathrm{Diversity} &= 1 - \frac{2}{B(B-1)} \sum_{1 \le m < n \le B} \mathrm{Id}\!\left(x^{(m)}, x^{(n)}\right).
\end{align}
Here, $\mathcal{A}$ is the set of amino acids observed in the generated set, $p(a)$ is the empirical frequency of amino acid $a$, and $B$ is the number of generated sequences being compared. Higher entropy and diversity indicate richer amino-acid usage and lower sequence-level collapse. We additionally report inner-TM, defined as the average pairwise TM-score between predicted structures in the generated set; lower inner-TM indicates that the sampled proteins are more structurally diverse.

\subsection{Language and Code Generation}
\label{app:language_experiments}

\paragraph{Open-dLLM} We use the Open-dLLM-0.5B code model \citep{opendllm2025} and train only the policy. Policy training uses FineCode \citep{opendllm2025}, the code corpus used by Open-dLLM. Each document is tokenized with the Open-dCoder-0.5B-no-PL tokenizer, terminated with an end-of-sequence token, and divided into chunks of at most 4,096 tokens without combining different documents. We evaluate pass@1 on HumanEval \citep{humaneval}, HumanEval+ \citep{evalplus}, MBPP \citep{mbpp}, and MBPP+ \citep{evalplus}.

\paragraph{LLaDA} We train the policy for 2,000 optimizer steps on a 100M-token mixture comprising 60\% FineWeb-Edu \citep{penedo2024fineweb}, 20\% OpenWebMath \citep{paster2024openwebmath}, and 20\% FineCode \citep{opendllm2025}, with a separate 5M-token validation split. Because LLaDA does not release its pretraining data, this mixture is used only for policy training. We evaluate LLaDA-8B \citep{nie2025llada} on HumanEval \citep{humaneval}, BBH \citep{suzgun2023challenging}, CoQA \citep{reddy2019coqa}, WMT16 English--Romanian \citep{bojar2016findings}, MBPP \citep{mbpp}, and XSum \citep{narayan2018xsum}. Generation length, denoising steps, and block length are respectively 512/512/512 for HumanEval; 256/256/256 for BBH and MBPP; and 64/64/64 for CoQA, WMT16, and XSum. In the reduced-step evaluation, the denoising budgets are divided by four while generation and block lengths are fixed at 512/128/512, 256/64/256, and 64/16/64, respectively. High-confidence ordering and the learned policy use identical generation settings. We evaluate using pass@1 task completion for HumanEval and MBPP, exact answer match for BBH, F1-score for CoQA, BLEU \citep{papineni2002bleu} for WMT16, and ROUGE-L \citep{lin2004rouge} for XSum.

For both open-dLLM and LLaDA tasks, we report the best result obtained from a hyperparameter sweep over inference settings (eg. token sampling temperature, position-sampling temperature, heuristic).

\section{Additional Policy Architecture Results}
\label[appendix]{app:transformer_policy}

\subsection{Transformer Policy Variant}

\begin{wraptable}{r}{0.58\textwidth}
    \vspace{-0.8em}
    \caption{Policy architecture comparison on Sudoku with deterministic decoding and $T{=}20$ steps.}
    \label{tab:transformer_policy}
    \centering
    \small
    \renewcommand{\arraystretch}{1.06}
    \resizebox{\linewidth}{!}{%
    \begin{tabular}{lc}
        \toprule
        \textbf{Policy architecture} & \textbf{Accuracy} \\
        \midrule
        Per-token MLP (scores + hidden) & \textbf{90.82\%} \\
        Score Transformer (scores only) & 89.84\% \\
        Score + Hidden Transformer (scores + hidden) & 90.23\% \\
        \bottomrule
    \end{tabular}}
    \vspace{-0.6em}
\end{wraptable}

\noindent In the policy-only setting, we also evaluate a transformer-based alternative to the per-token MLP. The score-only variant projects each confidence score to dimension 128 before a single transformer encoder layer. The score-and-hidden variant additionally incorporates the denoiser hidden states, allowing the policy to exchange information across positions before selecting an ordering.

The per-token MLP remains the strongest architecture in \cref{tab:transformer_policy}. Adding hidden states improves the transformer variant, but cross-position attention provides no further gain over the simpler MLP on Sudoku. This suggests that the denoiser hidden states already contain sufficient contextual information for the policy to make effective token-level decisions.

\section{Relation of Objective to ELBO Bounds}
\label[appendix]{app:elbo_relation}

We can derive our objective \cref{eq:order_obj} as an approximation to an ELBO bound. 

In particular, we make use of Proposition 3.2 from \citet{peng2025plannerawarepathlearning}. This gives an ELBO-style objective for masked diffusion decoding under an unmasking policy $q^\phi$. Rewritten in the notation of this paper, the proposition can be stated as follows.

\begin{proposition}[Proposition 3.2 of \citep{peng2025plannerawarepathlearning}]
\label{prop:papl_elbo}
Let $q^\phi(i \mid x_0, x_t)$ be a policy over masked positions at time $t$, and let
$p_{\theta, \phi}(x_0)$ denote the marginal probability of $x_0$ under the unmasking process
obtained by decoding according to this policy. For each time $t$, let
$r_t^{\phi}(\cdot \mid x_0)$ denote the distribution of the partially masked sequence $x_t$
under the policy-guided unmasking process.
Then
\begin{align}
    \log p_{\theta, \phi}(x_0)
    \ge
    \mathcal{E}_1^{\theta,\phi}(x_0)
    +
    \mathcal{E}_2^{\theta,\phi}(x_0),
\end{align}
where
\begin{align}
    \mathcal{E}_1^{\theta,\phi}(x_0)
    &=
    \mathbb{E}_{t \sim \mathcal{U}[0,1]}
    \left[
        c(\alpha_t)
        \mathbb{E}_{x_t \sim r_t^{\phi}(\cdot \mid x_0)}
        \left[
            \sum_{i=1}^{L}
            q^\phi(i \mid x_0, x_t)\,
            \mathrm{CE}(x_0^i,\mu^\theta(x_t)[i,\cdot])
        \right]
    \right],
\end{align}
and
\begin{align}
    \mathcal{E}_2^{\theta,\phi}(x_0)
    =
    -\mathbb{E}_{t \sim \mathcal{U}[0,1]}
    \left[
        c(\alpha_t)
        \mathbb{E}_{x_t \sim r_t^{\phi}(\cdot \mid x_0)}
        \left[
            \sum_{i=1}^{L}
            q^\phi(i \mid x_0, x_t)\,
            \log
            \frac{q^\phi(i \mid x_0, x_t)}
            {F_{\theta,\phi}(x_t, x_0^i, i)}
        \right]
    \right].
\end{align}
Here $c(\alpha_t)$ is some constant depending on the schedule, and $F_{\theta,\phi}(x_t, x_0^i, i)$ is an auxiliary distribution:
\begin{align}
    F_{\theta,\phi}(x_t, x_0^i, i)
    =
    \mathbb{E}_{z \sim \mu^\theta(x_t)}
    \left[
        q^\phi(i \mid z^{-i, x_0^i}, x_t)
    \right],
\end{align}
$z^{-i, x_0^i}$ is $z$ with token $i$ set to value $x_0^i$.
\end{proposition}

To simplify for our setting, since $x_0$ is sampled from the denoiser, we can consider the policy $q^\phi(i \mid x_0 \sim \mu^\theta(x_t), x_t)$ as being conditioned on only $x_t$, so that $q^\phi(i \mid x_t, x_0) = q^\phi(i \mid x_t)$. 
With this notation, we note that the first term $\mathcal{E}_1$ is very similar to the negative of the policy-weighted loss in \cref{eq:order_obj} evaluated along
policy-induced paths $r_t^{\phi}(\cdot \mid x_0)$ rather than the standard noising process $p(x_t \mid x_0)$.

The quantity $F_{\theta,\phi}(x_t, x_0^i, i)$ in the second term evaluates the policy probability of unmasking position $i$ to have token $x^i_0$ (marginalizing over possible denoiser randomness). Explicitly evaluating the $\mathcal{E}_2$ term requires multiple passes of the policy module.

From the above ELBO, if we make the following approximations, then we can recover our objective in \cref{eq:order_obj}:
\begin{itemize}
    \item sample $x_t$ from the noising distribution $p(x_t \mid x_0)$ rather than the policy-induced distribution $r_t^{\phi}(\cdot \mid x_0)$, since the former is tractable as simple. \citet{peng2025plannerawarepathlearning} also make this simplification.  
    \item we omit the second term $\mathcal{E}_2^{\theta,\phi}(x_0)$, which involves a term $F_{\theta,\phi}(x_t, x_0^i, i)$ that is difficult to compute. This term may be omitted when fixing the policy (such as when it is obtained from a heuristic). By contrast, we learn our policy, but argue that omitting $\mathcal{E}_2$ still yields a reasonable objective for training the policy, which we reinforce empirically. 
\end{itemize}

Given these approximations, we obtain our policy weighted objective as a surrogate of the policy-induced ELBO. We use this objective for both policy-only training (in \cref{eq:order_obj}), as well as joint training of the policy and denoiser (in \cref{eq:policy_aware_mdm_obj}).

\end{document}